\documentclass[conference]{IEEEtran}
\IEEEoverridecommandlockouts
\usepackage{cite}
\usepackage{amsmath,amssymb,amsfonts}
\usepackage{algorithmic}
\usepackage{graphicx}
\usepackage{textcomp}
\usepackage{xcolor}
\usepackage{hyperref}
\usepackage{listings}

\definecolor{codegreen}{rgb}{0,0.6,0}
\definecolor{codegray}{rgb}{0.5,0.5,0.5}
\definecolor{codepurple}{rgb}{0.58,0,0.82}
\definecolor{backcolour}{rgb}{0.95,0.95,0.92}

\lstdefinestyle{mystyle}{
    backgroundcolor=\color{backcolour},   
    commentstyle=\color{codegreen},
    keywordstyle=\color{magenta},
    numberstyle=\tiny\color{codegray},
    stringstyle=\color{codepurple},
    basicstyle=\ttfamily\footnotesize,
    breakatwhitespace=false,         
    breaklines=true,                 
    captionpos=b,                    
    keepspaces=true,                 
    numbers=left,                    
    numbersep=5pt,                  
    showspaces=false,                
    showstringspaces=false,
    showtabs=false,                  
    tabsize=2
}

\def\BibTeX{{\rm B\kern-.05em{\sc i\kern-.025em b}\kern-.08em
    T\kern-.1667em\lower.7ex\hbox{E}\kern-.125emX}}
\begin{document}
%
\title{Deep Optimal Timing Strategies for Time Series}

\author{\IEEEauthorblockN{Chen Pan,
Fan Zhou,
Xuanwei Hu, 
Xinxin Zhu, 
Wenxin Ning,
Zi Zhuang,
Siqiao Xue,\\
James Zhang,
and
Yunhua Hu
\IEEEauthorblockA{Ant Group, China}
\IEEEauthorblockA{\{bopu.pc, hanlian.zf, xuanwei.hxw, qinrui.zxx, wenxin.nwx, zhuangzi.zz, siqiao.xsq, james.z, wugou.hyh\}@antgroup.com}
}
}

\maketitle

\begin{abstract}
    Deciding the best future execution time is a critical task in many business activities while evolving time series forecasting, and optimal timing strategy provides such a solution, which is driven by observed data. This solution has plenty of valuable applications to reduce the operation costs. In this paper, we propose a mechanism that combines a probabilistic time series forecasting task and an optimal timing decision task as a first systematic attempt to tackle these practical problems with both solid theoretical foundation and real-world flexibility. Specifically, it generates the future paths of the underlying time series via probabilistic forecasting algorithms, which does not need a sophisticated mathematical dynamic model relying on strong prior knowledge as most other common practices. In order to find the optimal execution time, we formulate the decision task as an optimal stopping problem, and employ a recurrent neural network structure (RNN) to approximate the optimal times. Github repository: \url{github.com/ChenPopper/optimal_timing_TSF}
\end{abstract}


\section{Introduction}

While facing uncertain future, a decision maker usually relies on the prediction results. There is a class of commonly encountered problem that when the manager should take some important actions, which is involving a timing problem.
Among numerous scenarios that need to choose optimal time, let us consider two typical inventory management problems. The first one is an inventory manager trying to replenish the stocks within a given period, but the historical prices of the products in stores fluctuate over time. To minimize the cost, the products are expected to be purchased at their lowest prices. Another one we consider is about optimizing the positive lead time: There are many operations management problems involving making decisions beforehand, while random demands may come and go between placing order and receiving order. 

Recent years, decarbonization is one of the most discussed topics on the agenda of every environment summit or conference, and many countries have already implemented renewable energy into many industries, and renewable energy is influencing the electricity price because of the unstable supply of the low-output and intermittent generators \cite{renew_power22}. Uncertainty of the clean energy supply brings many new challenges to the industry and academics, thus time series forecasting is one of the critical technology~\cite{renew_challenges13}. There are some research results focus on the balance of supply and demand of energy to reduce the unstability of the power grid system, especially, the usage of demand response to reduce the impact of peak loads based on the optimal timing strategies, e.g., \cite{os_energey23,os_smart_grid_demand,xue_meta_2022,qu-2022-rltpp, xiao2023}. 

Lots of researchers have dived into this area and most of existing literature mainly focus on sequence optimal decision-making policies based on simulations \cite{BV2011} with prior information or reinforcement learning \cite{AJ2019}. In this paper, we consider a practical situation that the decision-maker can only place one order at a fixed time, but arrival time of orders can be negotiated, considering the prevailing high cross-boarder shipping cost. These products are perishable (say, fruits) and the company has no storage to hold them. Before products' arrivals, some customers would wait and some might leave. To maximize profit, these orders should arrive when the demands accumulate to the highest level. However, the customers tend to leave if the waiting time is too long. Our proposed approach can help decision makers forecast the demand and discover the optimal arrival times of products. 
Please note that this is our first attempt to tackle these practical decision problems and our proposed mechanism can be naturally extended to other situations considering the RNN structure we employ for approximation.

We formulate the management problems described above as a combination of probabilistic forecasting of time series and optimal stopping problem, which also has well-known applications on pricing options, e.g., American options, Bermudan options, and barrier options, etc \cite{Hull15}.
As a powerful tool for industrial production procedures, deep neural network has caught more and more attentions of researchers recently, which also motivates the application of deep neural networks in solving the optimal timing problem. In classical stochastic control theory, the optimal stopping problem is tackled through dynamic programming principle and is transformed into a partial differential equation of variational inequality type \cite{OSRules1978}. However, due to the curse of dimensionality, it is usually difficult to derive a stable numerical solution to the equation. 

The optimal stopping problem is a well-established mathematical model for option pricing with underlying stochastic processes following geometric Brownian motions or their refined analogues 
(e.g., stochastic volatility models and jump diffusion processes, etc.), and the classical simulation algorithm is based on Monte Carlo method \cite{Glasserman2004, Egloff2005, xue2021graphpp, wang2023enhancing}. However, in the real world, most of the underlying time series (e.g., solar and wind power) cannot be efficiently simulated. Even for the quotes of financial assets, one still needs to build sophisticated mathematical models, calibrated by the real-world data. Researchers conduct meticulous work in transforming physical probability to risk-neutral probability when estimating option prices under a complete market hypothesis which does not strictly hold usually \cite{ESG2016}. Hence, a strong prior knowledge to data is necessary for simulating future samples in the previous work, which makes the existing work inapplicable to most of the real-world problems. 

To address the difficulties discussed above, we combine the probabilistic forecasting technique and the optimal timing decision theory to better deal with real-world strategic decisions (see Figure \ref{fig_model_structure}). We also rely on the deep learning tools to handle large amount of real-world data. On the other hand, the DeepAR algorithm developed by Amazon is an efficient tool for probabilistic forecasting with proven success in practice with accessible open source code \cite{deepar2019, xue2023easytpp}, which is ideal for our forecasting task, facilitating the flexibility to generate Monte Carlo samples with various types of distributions. Then, with generated prediction samples, we are able to find the optimal timing strategies based on the optimization algorithm studied in \cite{BCJ2019,BCJW2021}. Furthermore, considering the non-anticipative property of stopping times, we employ an RNN structure to recursively estimate the indicator functions of the stopping times.

\section{Related Work}

To find an optimal timing strategy, we describe the problem as an optimization task, which is closely related to reinforcement learning, the basic method in the machine learning field alongside supervised and unsupervised learning. Recently, research on algorithms to solve optimal stopping problems under reinforcement learning framework has attracted great interest. Most of them focus on the pricing problem under Black-Scholes framework or its alternatives, which force the underlying time series to follow some predetermined dynamics \cite{BCJ2019,BCJW2021,DQL2021,DRLOP2021}, e.g., geometric Brownian motions, and diffusion processes with or without jumps. However, most time series could not be simply modeled as stochastic processes mentioned above, demanding better forecasting tools for business decisions, witnessed by recently advances, 
such as DeepAR algorithm \cite{deepar2019}, DeepVAR algorithm \cite{deepvar2020}, transformer for time series \cite{transformer2020, hao2023cl, shi2023language}, and temporal fusion transformers \cite{tft2020}.

Among the existing researches on solving optimal stopping problems within deep learning framework, \cite{BCJ2019, BCJW2021} developed an approach based on feed-forward neural networks to approximate optimal stopping times from Monte Carlo samples, demonstrating high accuracy in Bermudan max-call option pricing, American option pricing, optimal stopping problem for fractional Brownian motion, etc. In \cite{osgp22}, authors solve the optimal stopping problem via assuming the underlying stochastic process as a deep Gaussian process, and approximates the backward solving procedure. \cite{Hu2019} formulates optimal stopping problems in finance as a ranking response surface problem, employing deep neural networks to find the optimal policy, which is robust to the training data. \cite{CDBM2019} show a fast-converging algorithm of Q-learning with optimal asymptotic variance property via solving an optimal stopping problem in finance. For interpretation, \cite{CM2018} propose an interpretable method by constructing a binary tree to find the optimal policy from observed data.

\section{Problem Formulation}
In this section, we provide the formulation for the inventory management problems described in the introduction through minimizing the target cost function.
Formally, let us assume the current time is $T_0 > 0$ and there are $N \ge 1$ products to be purchased before the time $T_0 + T$ where $T > 0$. We have observed the historical unit price sequences $X_i(t), t\le T_0$ for each product $i, 1\le i \le N$, which form a $N$-dimensional time series. Then the total cost $v$ for replenishing the stocks can be expressed as
\begin{equation}
    v(t_1,\dots,t_N|T_0) = \sum_{i}\mathbf{E}\left[ a_i X_i(t_i)\big|X_i(s), s\le T_0\right],
\end{equation}
where $a_i$ is the given number of units of product $i$ to be purchased, $t_i\in[T_0+1, T_0+T]$ is the transaction time in the future, and the expectations are taken over the stochastic processes $\{X_i(t), t > T_0\}_{i=1}^N$. To the implementation, the stochastic processes are generated sample paths from a prediction model, and the conditional expectations are substituted by sample means. That is, with a given (deep neural) probabilistic forecasting model $F(\cdot|\theta), \theta$ is the parameters, the predicted sample path $j\in \{1, \dots, J\}$ for product $i$ is 
\begin{equation}
    {X}_{i}^{(j)}(t) \sim F(\{X_i(s); s\le T_0\}, t| \theta_i),\quad T_0+1 \le t \le T_0+T,
\end{equation}
and the approximated total cost is 
\begin{equation}
    \frac{1}{J}\sum_{i} \sum_{j=1}^J a_i X_i^{(j)}(t_i).
\end{equation}

The minimum cost can be expressed as an optimal stopping problem as follows
\begin{equation}\label{main_problem}
    V(T_0) = \min_{\tau_i\in \mathcal{T}}\sum_{i}\mathbf{E}\left[ a_i X_i(\tau_i)\big|\mathcal{F}_{T_0}\right],
\end{equation}
where $\mathcal{F}_t=\{X_i(s), s\le t; 1\le i \le N\}$ is the history information that can be obtained up to time $t$ including all the product prices observed till time $t$, and $\mathcal{T}$ is the set of all stopping times taking values in $[T_0+1, T_0 + T]$. 

To better formulate our problem, we assume that the available transaction time in the future are $t_i \in \{T_0+1, \dots, T_0 + T\}$. 
Now problem (\ref{main_problem}) takes similar form as pricing a Bermudan option \cite{Hull15} with our special settings.

\section{Methods}


\begin{figure*}[ht]  
    \centering
    \includegraphics[width=0.8\linewidth]{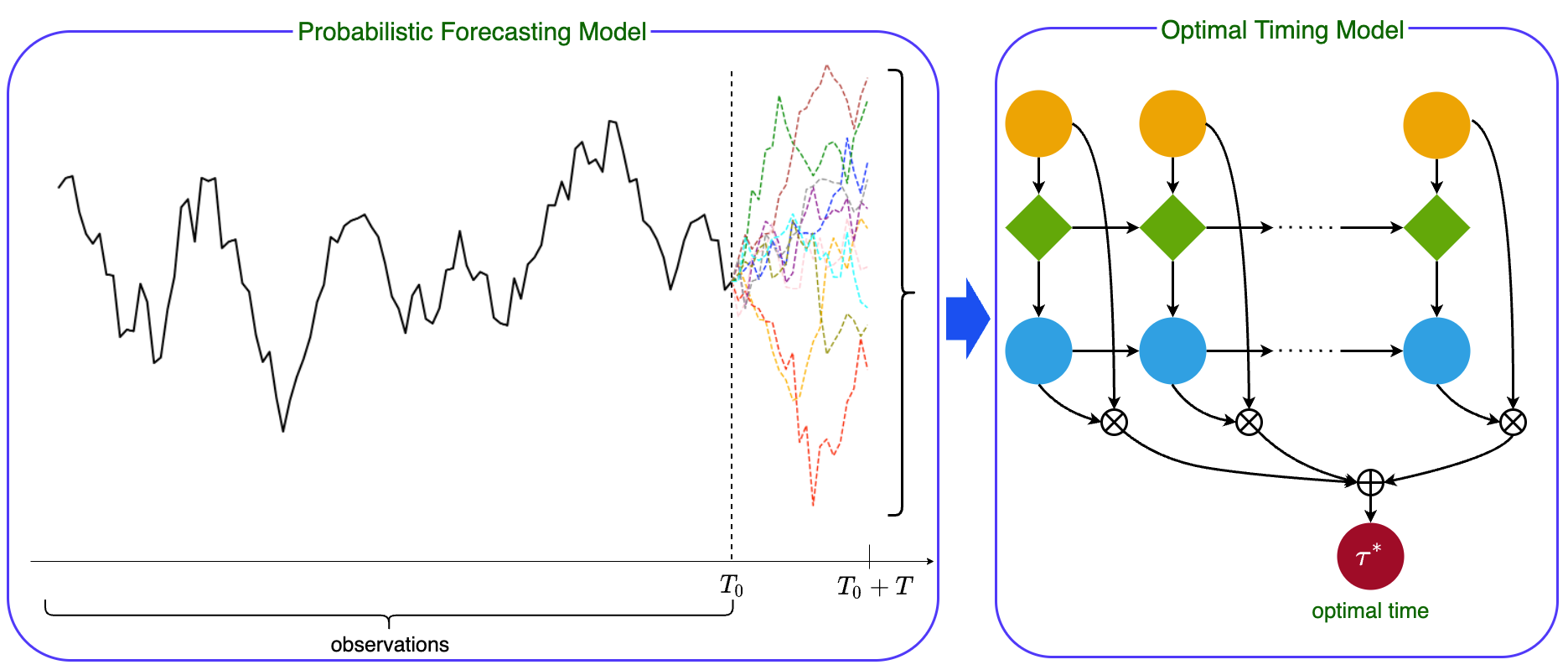} 
    \caption{The model structure: Feeding the history data $X_i(0:T_0)$ to the probabilistic forecasting model, which provides the prediction sample paths, and outputting the approximated stopping times $\tau^*$ through the optimal timing strategy model. }
    \label{fig_model_structure}
\end{figure*}

We provide the detailed algorithm in this section including a forecasting module and a decision module illustrated in Figure \ref{fig_model_structure}. 

Since the optimal timing strategy relies on the future prices of the products, i.e., $X_i(T_0+t), 1 \le t \le T$, we need to predict them according to the history information as a necessary step towards optimal timing. Therefore, we apply the probabilistic forecasting technique to generate the samples of the future prices, which avoids building mathematical models based on strong prior knowledge of the history data. 

The decision module mainly refers to \cite{BCJ2019,BCJW2021}, but here we optimize the stopping times for individual items separately,
which is more appropriate for our practical inventory problems. We also employ an RNN structure to update the approximated stopping times.

We rewrite equation (\ref{main_problem}) as
\begin{equation}\label{main_problem_2}
    V(T_0) = \min_{\tau_i\in \mathcal{T}}\mathbf{E}\left[g(X_1(\tau_1), \dots, X_N(\tau_N))\big|\mathcal{F}_{T_0}\right],
\end{equation}
where $g:\mathbf{R}^{N}\to \mathbf{R}$ is a predetermined function, we set $g(x_1,\dots, x_N) = \sum_{i=1}^N a_i x_i$, and for given stopping times $\tau_i, i=1, \dots, N$, one can express
\begin{equation}
    \begin{aligned}
    &\, g(X_1(\tau_1), \dots, X_N(\tau_N))\\
 = & \sum_{k_i} \delta_{\{\tau_i=k_i, i=1,\dots,N\}} g(X_1(k_1), \dots, X_N(k_N)). 
    \end{aligned}
\end{equation}


We denote $\mathbf{X}_t = (X_1(t), \dots, X_N(t))^\top$, and $\mathbf{X}_{m:n} = \mathbf{X}_{m}, \dots, \mathbf{X}_{n}$ for $m < n$. 

Referring to \cite{BCJ2019, BCJW2021}, the stopping time can be decomposed into a summation of indicator functions, which is
\begin{equation}
    \tau = \sum_{t=1}^{T} (T_0 + t) \mathbf{\delta}_{\{\tau=T_0+t\}},
\end{equation}
where the indicator functions $\delta_{\{\tau=T_0+t\}}$ are random variables, which are generated from the underlying samples $\{\mathbf{X}_{T_0+s};1\le s \le t\}$, thus they can be approximated by some functions $h_t$ (e.g., a neural network) as follows
\begin{equation}
    \mathbf{\delta}_{\{\tau=T_0+t\}} \approx D_{t}(\mathbf{X}_{T_0+1:T_0+t}), \quad 1\le t \le T.
\end{equation}
Let us focus on the approximated function  $D_t: \mathbf{R}^{N\times t}\to (0, 1)$, and it has an expression of
\begin{equation}\label{approx_equation}
    D_t = \max\{h_t(\mathbf{X}_{T_0+1:T_0+t}), t+1-T\}\left(1 - \sum_{s=1}^{t-1} D_s\right),
\end{equation}
where the functions $h_t: \mathbf{R}^{N\times t}\to (0, 1)$  is considered as another approximation of the indicator function $\delta$, which is characterized by a multi-layer feed-forward neural networks with Sigmoid activation functions. For detailed proof of equation \eqref{approx_equation}, one can refer to Section 2 of \cite{BCJW2021}, which involves basic mathematical analysis tools. It is worth to mention that the direct result of expression \eqref{approx_equation} is $\sum_{t=1}^T D_t = 1$.

With equation (\ref{approx_equation}) in hand, we are able to utilize the deep learning technology to solve the optimal stopping problem (\ref{main_problem_2}). For simplicity, we assume that $a_i\equiv 1$,
and then the optimal stopping problem takes the form of
\begin{equation}
    V(T_0) = \min_{\tau_i \in \mathcal{T}}\sum_{i}V_i(\tau_i|\mathcal{F}_{T_0}),
\end{equation}
where
\begin{align}
     V_i(\tau_i|\mathcal{F}_{T_0}) &= \mathbf{E}\left[\sum_{t=1}^{T}\delta_{\{\tau_i = T_0 + t\}}X_i(T_0+t)\right] \\
     &\approx \mathbf{E}\left[\sum_{t=1}^{T}D_{i,t}X_i(T_0+t)\right],
\end{align}
and the optimization target is to minimize the functional
\begin{equation}\label{target_functional}
    \sum_i \mathbf{E}\left[\sum_{t=1}^{T}D_{i,t}X_i(T_0+t)\right].
\end{equation}

\begin{figure}[t]
    \centering
    \includegraphics[width=0.9\columnwidth]{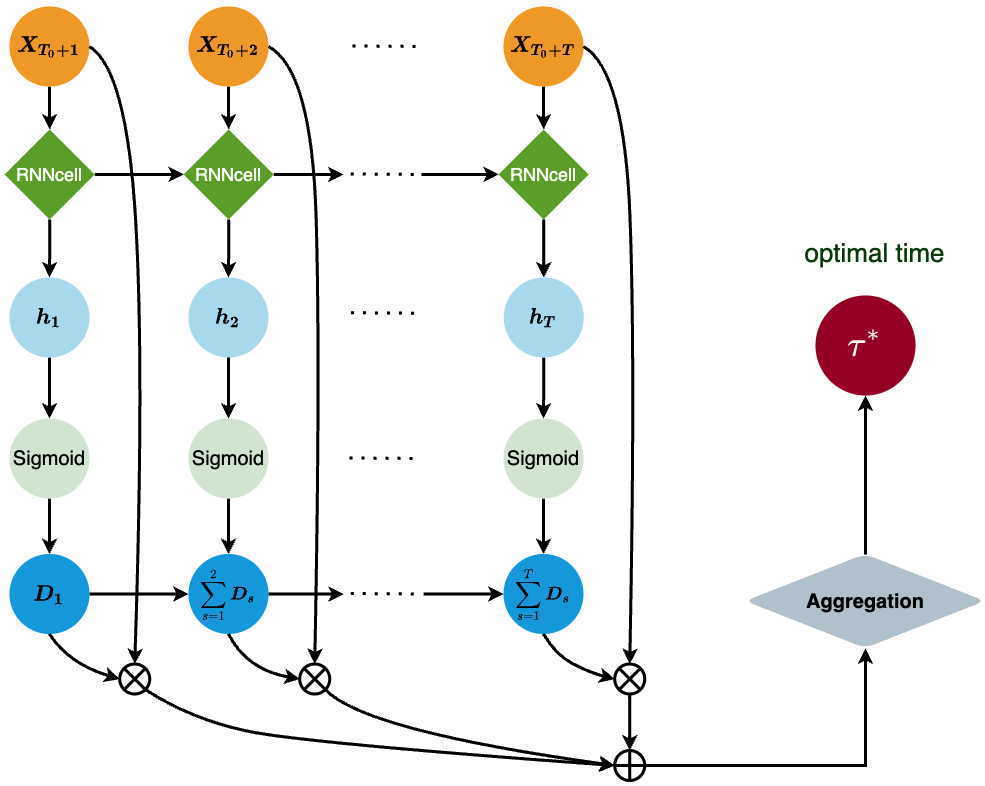} 
    \caption{The decision model has an RNN structure that incorporates the Markov property of the time series, and as an approximation of stopping time, $D_t$ is non-anticipative, i.e., it does not contain any future information.}
    \label{fig_rnn}
\end{figure}

Noticing that the approximations $D_t$ of the indicator functions of stopping times does not contain the future information (the non-anticipative property), we use an RNN structure to recursively estimate them. Specifically, we model the functions $h_t$ as outputs of feed-forward linear layers with Tanh activation functions for hidden layers and Sigmoid activation functions for output layers, since the functions are restricted to the range between 0 and 1. According to definitions of functions $h_t$, they also depend on the predicted sample paths up to time $t$, and thus an RNN structure is adapted to encode the inputs of those sample paths (see Figure \ref{fig_rnn}). 
To emphasize the parameters of the network, we rewrite the function as $h(t;\theta)$, where $\theta$ denotes the weights of the networks, and we use the stochastic gradient descent algorithm to train the neural network estimating the optimal $\hat{\theta}$. 

To calculate the target functional in equation (\ref{target_functional}), we estimate the values on sample paths and then take the means of them to approximate the expectations, which relies on the probabilistic forecasting results. Specifically, suppose the sample path is indexed by $j$ and $1 \le j \le J$, for product $i$, the expectations can be approximated as 
\begin{equation}
    \frac{1}{J} \sum_{j=1}^J \sum_{t=1}^T D_{i,t}^j(\theta)X^j_i(T_0+t),
\end{equation}
where $D_{i,t}^j(\theta)$ is rewritten to emphasize its dependence on path $j$ and parameter $\theta$.

Then for each sample path $j$, the estimated optimal stopping time for the $i$-th asset is given as
\begin{equation}
\hat{\tau}_i^j(\hat{\theta}) = \min\left\{1\le t\le T \bigg| \sum_{s=1}^t D_{i,s}^{j}(\hat{\theta}) \ge 1 - D_{i,t}^{j}(\hat{\theta})\right\}.
\end{equation}

The decision strategy is given by the mode of each $\hat{\tau}_i$, i.e.,
\begin{equation}
    {\tau}^*_i \approx T_0 + \arg\max \left\{1\le t\le T: \sum_{j=1}^J\delta_{\{\hat{\tau}_i^j=t\}}\right\}.
\end{equation}


\section{Empirical Results}

To perform the forecasting task, we employ three probabilistic forecasting algorithms, including DeepAR, DeepVAR, and Transformer for time series. Then we apply RNN-type modules (especially, GRU \cite{GRU14}) to deal with the decision task. The complete source code will be released to the public later, while the code for decision module can be referred to github repository.  

\paragraph{DATASETS} We test our model on two real-world datasets. The first one is recent five-year historical daily prices of twelve stocks from Nasdaq stock exchange, and the second dataset is the eight exchange rates from the public dataset of the exchange rates, which contains 27-year history exchange rates (from 1990 to 2016). 

Noticing that the stock prices vary from each other drastically and the highest stock price would cause the bias in the result, we normalize the stock prices to 1 on the last date of training set, whose rationality is to separate the budget equally on individual products. 

For stock prices, we set the prediction length by $T=20$ days and train the forecasting models in the first four years (2016/9/1 - 2020/10/31). For the exchange rate dataset, we set the prediction length by $T=5$ days, and train the forecasting models from 2008/1/1 to 2008/12/31.

\paragraph{BASELINE} The literature on the combination of forecasting and decision on real-world data is rare, and we therefore consider the 
natural baseline, which is the decision strategy (i.e., the optimal execution time) is chosen directly according to the predicted lowest prices for stocks and exchange rates. 

\begin{figure}[t]
    \centering
    \includegraphics[width=0.9\columnwidth]{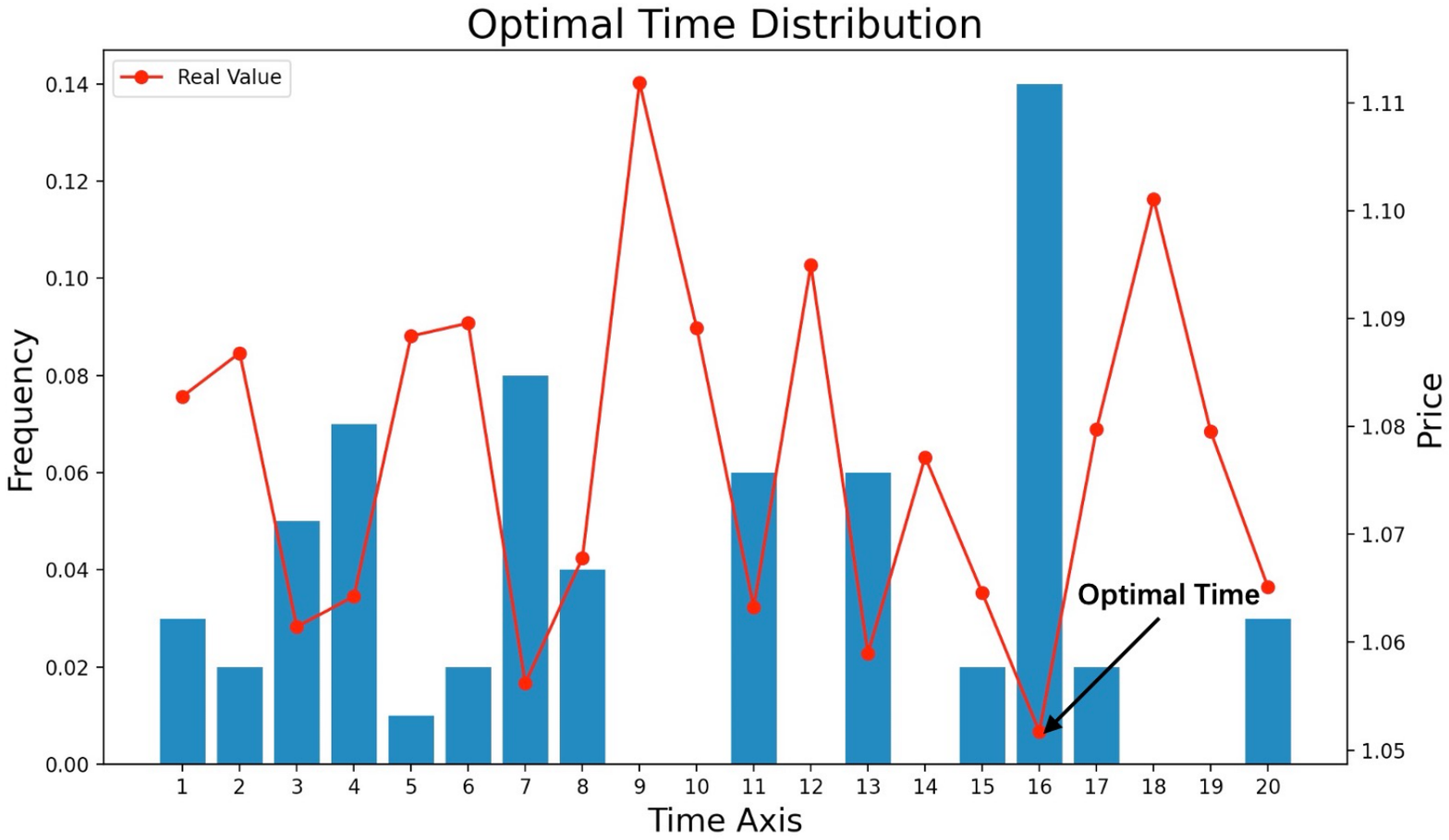} 
    \caption{An example of distribution for the estimated optimal time and one actual sample path of stock price (scaled).}
    \label{fig_mode}
\end{figure}

We calculate the optimal cost according to the decision made on prediction result and the decision model respectively. Then, we use accuracy (i.e., $1 - MAPE$) with respect to the actual minimum cost as the metric of performance, and compare the accuracy between the mode of optimal times on the predicted sample paths and the baseline. Figure \ref{fig_mode} shows the probability distribution of an exemplar estimated optimal execution time and the (future) actual price sample path, in which one can see the time with the highest frequency is when the price reaches the lowest level. 

\paragraph{RESULTS} For stock prices, we calculate the MAPEs for decision result from 2020/12/1 to 2021/1/29, and find our algorithm (DeepAR+OSD, DeepVAR+OSD, and Transformer+OSD) has significant advantage (see Table \ref{table_stock}, the $p$-value for $t$-test is less than $0.05$) over the baseline. 
For the other public dataset of exchange rates, we calculate the accuracy for decision result from 2009/2/15 to 2009/3/31, and our result also demonstrates superior performance over DeepAR forecasting algorithm, and significant advantage against the other two forecasting algorithms (see Table \ref{table_exchange}).
\begin{table}
    \begin{minipage}{3.5in}
    \centering
    \begin{tabular}{l|l|l}
        \hline
         & Accuracy & Advantage \\
        \hline\hline
        DeepAR &  0.9557 &  \\
        DeepAR+OSD & 0.9733 & 0.0177 (0.0000)\footnote{p-value for $t$-test} \\
        \hline
        DeepVAR & 0.9578 &  \\
        DeepVAR+OSD & 0.9725 & 0.0166 (0.0000)\\
        \hline
        Transformer & 0.9544 & \\
        Transformer+OSD & 0.9735 & 0.0191 (0.0005)\\
        \hline
    \end{tabular}
\end{minipage}
    \caption{Performance on Nasdaq dataset (2020/12/1 -- 2021/1/29)}
    \label{table_stock}
\end{table}
\begin{table}
    \begin{minipage}{3.5in}
    \centering
    \begin{tabular}{l|l|l}
        \hline
         & Accuracy & Advantage \\
        \hline\hline
        DeepAR &  0.9929  & \\
        DeepAR+OSD & 0.9949 & 0.0020 (0.0010) \\
        \hline
        DeepVAR & 0.9953 & \\
        DeepVAR+OSD & 0.9960 & 0.0007 (0.0149)\\
        \hline
        Transformer & 0.9949 & \\
        Transformer+OSD & 0.9955 & 0.0006 (0.0420)\\
        \hline
    \end{tabular}
    \vspace{0.1in}
    \end{minipage}
    \caption{Performance on exchange rate dataset (2009/2/15 -- 2009/3/31)}
    \label{table_exchange}
\end{table}

\section{Conclusion}
In this paper, we have proposed a mechanism that combines a probabilistic forecasting task and an optimal timing decision task to solve a class of timing problems in several real-world scenarios. For the probabilistic forecasting task we employed DeepAR, DeepVAR, and Transformer algorithms, while for the decision task we constructed an RNN-type deep neural networks to optimize the cost functional. The empirical results demonstrated significant advantages of our algorithm in reducing the total cost. As a future work, we would consider an end-to-end model for combining forecasting task and decision task.

\end{document}